\documentclass[runningheads]{llncs}
\usepackage{graphicx}

\usepackage{tikz}
\usepackage{comment}
\usepackage{amsmath,amssymb} 
\usepackage{color}

\usepackage{graphicx}
\usepackage{bbm}
\usepackage{bm}
\usepackage{xspace}
\usepackage[utf8]{inputenc} 
\usepackage{url}            
\usepackage{booktabs}       
\usepackage{amsfonts}       
\usepackage{nicefrac}       
\usepackage{microtype}      
\usepackage{algorithm}
\usepackage{algorithmic}
\usepackage{array}
\usepackage{multirow}
\usepackage{enumitem}
\usepackage{makecell}
\usepackage{gensymb}         
\usepackage{mathtools}
\usepackage{dblfloatfix}
\usepackage{pifont}
\usepackage[permil]{overpic}

\usepackage{titletoc,tocloft}

\makeatletter
\DeclareRobustCommand\onedot{\futurelet\@let@token\@onedot}
\def\@onedot{\ifx\@let@token.\else.\null\fi\xspace}

\def\ie{\emph{i.e}\onedot}

\def\etal{\emph{et al}\onedot}
\makeatother

\usepackage{xcolor}
\usepackage{ifthen}
\usepackage{textcomp}
\definecolor{red}{rgb}{1,0,0}
\definecolor{slateblue}{rgb}{0.7,0.35,0.9}
\definecolor{green}{rgb}{0,1,0}
\definecolor{mahogany}{rgb}{0.75, 0.25, 0.0}
\definecolor{purple}{rgb}{0.6, 0, 0.6}
\definecolor{darkpurple}{rgb}{0.3, 0, 0.3}
\definecolor{darkgreen}{rgb}{0, 0.4, 0}
\definecolor{frenchblue}{rgb}{0.0, 0.45, 0.73}
\definecolor{blue}{rgb}{0,0,1}
\definecolor{goldenrod}{rgb}{0.65, 0.45, 0.03}
\definecolor{gray}{rgb}{0.5,0.5,0.5}
\definecolor{gold}{rgb}{1.0, 0.874, 0}
\definecolor{silver}{rgb}{0.67,0.67,0.67}
\definecolor{brown}{rgb}{0.8, 0.678, 0.4}

\usepackage[accsupp]{axessibility}  

\usepackage[width=122mm,left=12mm,paperwidth=146mm,height=193mm,top=12mm,paperheight=217mm]{geometry}

\usepackage{cite}
\usepackage[pagebackref,breaklinks,colorlinks]{hyperref}
\usepackage[capitalize]{cleveref}

\graphicspath{ {./figures/} }

\begin{document}
\pagestyle{headings}
\mainmatter
\def\ECCVSubNumber{4740}  

\title{Mask2Hand: Learning to Predict the 3D Hand Pose and Shape from Shadow} 

\titlerunning{Mask2Hand: Learning to Predict the 3D Hand Pose and Shape from Shadow}
%
\author{Li-Jen Chang \and
Yu-Cheng Liao \and
Chia-Hui Lin \and 
Hwann-Tzong Chen }
\authorrunning{L.-J. Chang et al.}
%
\institute{National Tsing Hua University}
\maketitle

\begin{abstract}
We present a self-trainable method, {\em Mask2Hand}, which learns to solve the challenging task of predicting 3D hand pose and shape from a 2D binary mask of hand silhouette/shadow without additional manually-annotated data. Given the intrinsic camera parameters and the parametric hand model in the camera space, we adopt the differentiable rendering technique to project 3D estimations onto the 2D binary silhouette space. By applying a tailored combination of losses between the rendered silhouette and the input binary mask, we are able to integrate the self-guidance mechanism into our end-to-end optimization process for constraining global mesh registration and hand pose estimation. The experiments show that our method, which takes a single binary mask as the input, can achieve comparable prediction accuracy on both unaligned and aligned settings as state-of-the-art methods that require RGB or depth inputs.
Our code is available at \url{https://github.com/lijenchang/Mask2Hand}.
\end{abstract}

\begin{figure}[t]
    \centering
    \includegraphics[width=\textwidth]{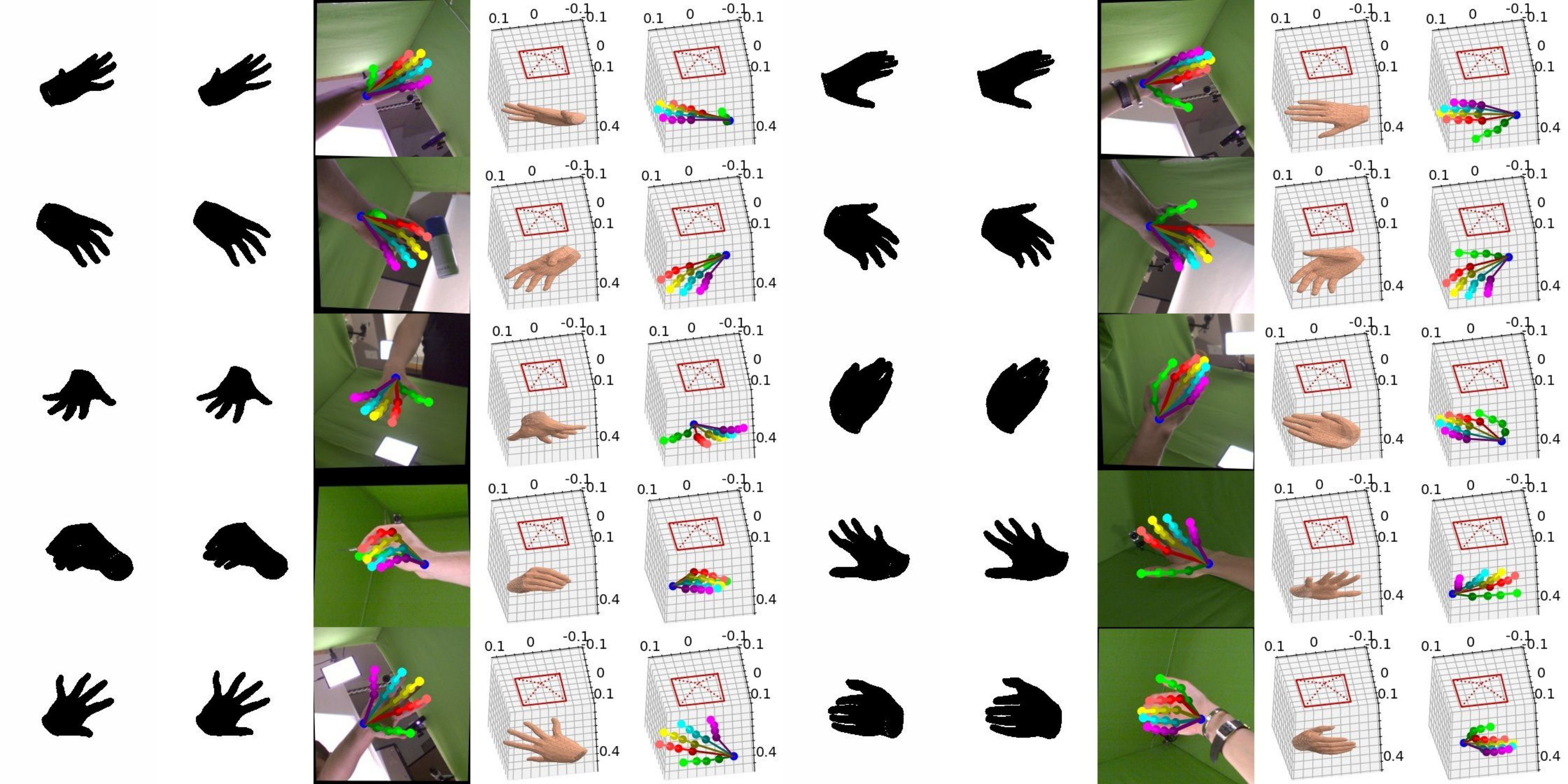}
    \caption{
    The proposed method, Mask2Hand, learns to solve the challenging task of predicting the 3D hand pose and shape from a 2D binary mask of hand silhouette. Each row in the left or right half of the figure illustrates an example result of performing our method on real data. The first column shows the input of 2D binary mask. The second column shows the rendered mask using the predicted hand shape. The third column shows the predicted 2D skeleton depicted in the image space. Note that the color images are simply for visualization; our method does not use any RGB data. The last two columns show the predicted mesh and 3D skeleton.
    } 
    \label{fig:teaser}
\end{figure}

\section{Introduction}

More than hundreds of years ago the historic art form of hand shadow puppetry had already been documented in many ancient countries. Since then, making hand shadow puppets has been a low-cost household entertainment activity that parents can play with their children.
As far as human perception is concerned, it is interesting that humans can recognize the animals and characters portrayed by the hands and fingers from merely their projected shadows. It is also remarkable that people are able to develop and create different ways of making hand shadow puppets, which may involve the generative capability of imagining some target shape and adjusting the hands and fingers to match the imaginary shape with the shadow.
Inspired by the discriminative and generative nature in the visual process of doing hand shadow puppets, we aim to explore the possibility of learning to estimate the 3D hand pose and shape from an input hand silhouette, through a similar trial-and-error manner of learning to do hand shadow puppets.

Our main goal in this endeavor is to train a renderer-equipped deep network that takes a binary mask of a hand's silhouette as the input and predicts the corresponding 3D hand shape that produces a silhouette very similar to the input. (\cref{fig:teaser} shows several example results obtained using our model.) To build such a model, we employ MANO \cite{RomeroTB17} as the 3D hand-shape renderer in our deep network. We propose a new network, called Mask2Hand, which consists of an encoder, a MANO layer, a differentiable render, and the refinement module. \cref{fig:model} shows an overview of the proposed Mask2Hand network. Our network takes a binary mask as the input and uses the encoder to generate the required parameters for the MANO layer to render the hand mesh. 
The encoder learns to predict the pose-related principal components of MANO and the 3D global transformation of the hand from only a single input binary mask.
Based on the principal components and global transformation, the MANO layer reconstructs the 3D hand mesh and the hand joint positions. We then use a differentiable renderer to generate the 2D projected silhouette from the hand mesh. Finally, the refinement module compares the projected silhouette with the input binary mask to refine the entire network. Three groups of loss functions are tailored for training the aforementioned modules, including the pose loss, silhouette loss, and mesh loss. The loss functions and the operations involved in the modules are all differentiable---we are able to train the entire network from end to end. 

As this work aims at addressing a rather new and challenging task of predicting the 3D hand pose from a single 2D binary mask, it is quite encouraging that the proposed Mask2Hand network can achieve comparable prediction accuracy on 3D hand pose and shape estimation as state-of-the-art methods, although we only need a single binary input image while theirs require an RGB image or a depth map. Note that our model can be trained under supervision with manually annotated joint positions or in a fully self-supervised manner without human annotations. Furthermore, since our method learns to predict the global transformation of the hand, its evaluation performance under the unaligned setting of hand pose estimation does not degrade as much as other methods do. Such a property is particularly favorable as in real applications it is not allowed to align the prediction with the unknown ground truth.

In addition to the motivation of hand shadow puppets, there are some other advantages and real-world applications of working on binary silhouettes. First, in contrast to RGB input, binary images can be obtained with relatively inexpensive sensors, such as low-resolution range sensors and low-resolution infrared sensors. They are especially useful when the lighting condition is poor and not much color information is available. Second, a line of research~\cite{LiXXHYZ17,MaLHHU0Y19,VarshneySMV17} in the hardware domain explores the usage of ambient light as a ubiquitous sensing medium. Light sensors, such as photodiodes or solar cells, measure and track the shadow blockage that a user's hands cause with low cost and ultra-low power consumption, which have the potential to be widely used in IoT devices for shadow-based hand pose estimation systems. Finally, binary input is more robust to adversarial attacks of imperceptible pixel perturbation. From the literature on adversarial machine learning~\cite{AkhtarMKS21}, we know that invisible image manipulation can seriously spoil the functionality of DNNs. However, this common type of attack is not directly applicable to the binary setting since small perturbation in pixel values is impossible with binary images. Also, pixel inversion in binary silhouettes can be easily identified and addressed.



\section{Related Work}

One of the research endeavors inspiring us is {\em Shadow Theatre}~\cite{WonL16}, which aims to reproduce computationally the shadow puppet theatre---an ancient shadow-oriented form of performance art---just like how we are motivated by hand shadow puppetry. Won and Lee ~\cite{WonL16} develop an algorithm to incorporate several features and heuristic rules that are related to shadow generation using human bodies. The proposed method optimizes the poses of human bodies to match the given 2D target shapes. The underlying nonlinear high-dimensional optimization is solved using the proposed heuristic strategies so that a plausible solution can be found with reasonable computing resources.
Nevertheless, the computation still requires tens to hundreds of minutes to complete on GPU.
\subsection{3D Hand Pose and Shape Estimation}
Various methods have been proposed to address the problem of 3D hand pose and shape estimation. Generally, the techniques that are used to solve 3D human pose estimation can also be applied to 3D hand pose estimation. Based on the input format, we consider the following four categories of methods on 3D hand pose and shape estimation: RGB-based, depth-based, event-camera-based, and silhouette-based methods. Among them, silhouette-based methods are most closely related to this work, where we seek to generate the 3D hand pose and mesh from merely a single binary-valued image of hand silhouette. 

Despite the discrepancy in their input type, many of the methods take 2D joint detection as the basis. Such an approach is promising and many existing deep-learning-based techniques can be employed. The following is a brief review of recent methods in the four categories.

\paragraph{RGB-based Methods.}
Lin \etal propose a method called Mesh Transformer (METRO) \cite{LinWL21} that explores non-local relationships between vertices and joints to reconstruct 3D human pose from a single RGB image without using parametric mesh models. This method can also be extended to hand mesh reconstruction.
Chen \etal~\cite{ChenLMCWCGWZ21} propose Camera-space Mesh Recovery (CMR) that generates 3D meshes from a single RGB image by extracting 2D cues including joint landmarks and silhouette.
Chen \etal~\cite{ChenTKBZZCY21} present a self-supervised 3D hand reconstruction network called $\text{S}^{2}\text{HAND}$ that estimates 3D poses from 2D detected keypoints.
Liu \etal~\cite{LiuJXLW21} propose a semi-supervised 3D hand-object pose estimation method leveraging spatial-temporal consistency in videos to obtain pseudo-labels for self-training.
Li \etal~\cite{LiGS21} propose a hand pose estimation method by using multi-task learning to categorize joints into groups so that different features can be learned to recover the 3D joint locations in a group-wise manner.

\paragraph{Depth-based Methods.}
Xiong \etal ~\cite{XiongZXCYZY19} propose the Anchor-to-Joint Regression Network (A2J) that exploits informative anchor points to enhance the generalization ability of 3D pose estimation.
Fang \etal ~\cite{FangLLXK20} present a dense-prediction based 3D hand pose estimation method called JGR-P2O that uses a pixel-to-offset prediction network and a joint graph reasoning module to enable end-to-end training and improve computational efficiency.

\paragraph{Asynchronous-event-based Methods.} 
Event cameras can provide high-dynamic-range and high-temporal-resolution input data, which are useful for many vision tasks. However, event cameras are more expensive and less popular than standard RGB cameras, and therefore not many methods have been proposed to use event cameras for 3D hand pose estimation. A very recent method proposed by Rudnev \etal \cite{RudnevGWSMET21}, which is called EventHands, manages to estimate 3D hand poses using a real-time asynchronous event camera, and it outperforms color-based and depth-based methods.

\paragraph{Silhouette-based Methods.}
Silhouette-based methods have not yet been widely applied to hand/body pose estimation. The task to be solved is more challenging than those in other settings of RGB-based and depth-based methods. The difficulty comes from the lack of information in the input data, where shading, color, and depth of the 3D objects are absent in the binary-valued images. Lee \etal present the Silhouette-Net ~\cite{LeeLCI19} method that generates 3D hand poses based on binary-valued silhouettes and can achieve a performance similar to or better than depth-based methods if multiview silhouettes are provided. However, their method requires the guidance of depth maps during the training stage.
In contrast, our method does not require additional depth information during training, and it only needs a single-view 2D binary mask of the hand as the input. Despite the challenging setting, estimating the 3D hand pose and mesh from limited binary information in the input is our major goal in this work.

\subsection{3D Hand Models}
\begin{figure*}[t]
    \centering
    \includegraphics[width=\textwidth]{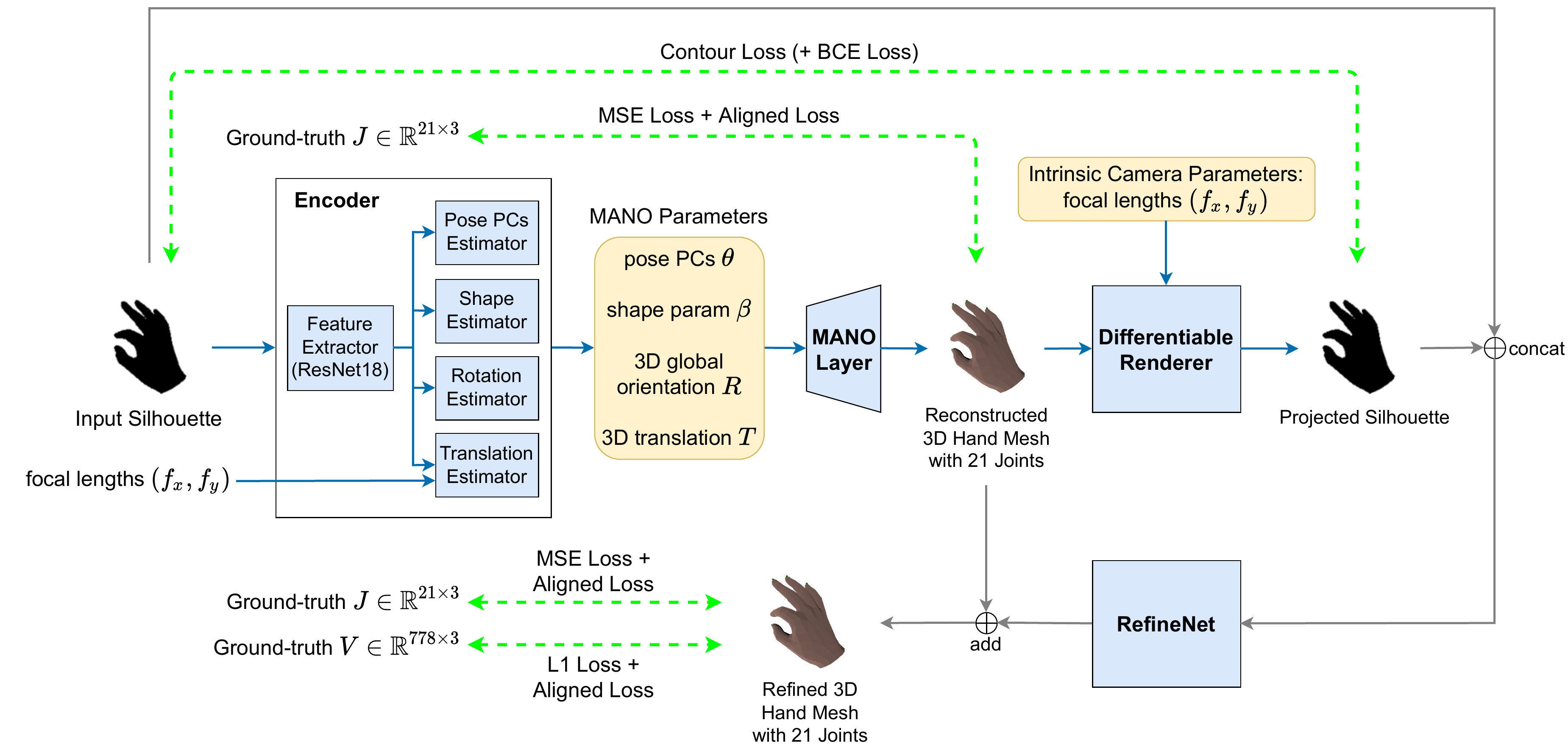}
    \vspace{-1em}
    \caption{The Mask2Hand network comprises the Encoder, MANO Layer, Differentiable Renderer, and RefineNet. 
   The Encoder takes a binary mask as the input and generates the required parameters for the MANO Layer; the parameters include the shape, the six or forty-five pose-related principal components of MANO, the 3D global orientation, and translation of the hand. Based on these parameters, the MANO Layer reconstructs the 3D hand mesh (represented by 778 vertices) with the 21 positions of the hand joints. The Differentiable Renderer generates the 2D projected silhouette of the hand mesh, and the RefineNet can compare the projection with the input binary mask to refine the entire model. We design different loss functions for different modules in our model. The green dashed lines indicate what losses are applied to the associated modules. Please refer to the text for the detailed formulations of the loss functions.  
    }
    \label{fig:model}
\end{figure*}

Several hand mesh models have been proposed, such as MANO \cite{RomeroTB17}, Sphere-Meshes \cite{TkachPT16}, and Convex Parts \cite{TzionasBSAPG16}. With the help of these models, hand shape generation and pose estimation can be processed via parametric methods.
Typically, a hand model like MANO is designed to parameterize a triangle mesh into pose, shape, and rotation parameters. MANO has been widely used by hand pose estimation methods \cite{BoukhaymaBT19, BaekKK19, ChoutasPBTB20, CoronaPAMR20, HassonTBLPS20, MoonL20, SeeberPPO21, Wang0BSSQOCT20, ZhangLMZZ19, ZhouHXHT020} to train their deep networks; the deep networks learn to predict the hand shape in the target image by regressing to the principal components of the MANO model. 
As shown in \Cref{fig:model}, our method also uses the MANO model to construct the target hand shape. By learning to optimize the parameters of MANO meshes, our method achieves the goal of generating the 3D hand mesh from a binary-valued silhouette image.



\section{Our Approach} \label{sec:approach}

To predict the hand pose and shape from a binary mask, we propose a two-phase end-to-end trainable network, \emph{Mask2Hand}, that aims to reconstruct 3D hand meshes in the camera space. \cref{fig:model} illustrates the overall pipeline of our method. In what follows, we detail each module of the pipeline and describe how we formulate the loss functions. 
                                
\subsection{Model Architecture} \label{ssec:model}

The core modules of our model include an encoder for estimating the MANO parameters, a MANO layer that reconstructs the 3D hand mesh, a differentiable renderer for optimizing the hand pose, and a RefineNet improving the accuracy. 

\paragraph{Encoder.} Given a hand silhouette image presented as a binary mask, the encoder first uses a ResNet18~\cite{HeZRS16} backbone to extract features from the input. Then, these features are fed into four different estimators to encode 3D information into the PCA space of MANO~\cite{RomeroTB17} and predict the extrinsic camera parameters. More specifically, the output of the encoder consists of the first six or forty-five hand-pose-related principal components $\theta \in \mathbb{R}^{6}$ or $\mathbb{R}^{45}$ of MANO, shape parameters $\beta \in \mathbb{R}^{10}$, 3D rotation $R \in \mathbb{R}^3$ in axis-angle representation, and 3D translation $T \in \mathbb{R}^3$.

\paragraph{MANO Layer.} The 3D hand shape is represented by a triangle mesh that contains a vertex set $V \in \mathbb{R}^{778 \times 3}$ and fixed faces $F$ indicating the connection between them. In our network, the hand mesh and joints $J \in \mathbb{R}^{21 \times 3}$ are reconstructed using MANO ~\cite{RomeroTB17}. Being a low-dimensional parametric hand model, MANO can generate the required hand mesh from parameters including the hand-pose-related principal components $\theta$, shape $\beta \in \mathbb{R}^{10}$, rotation $R$, and translation $T$. 

\paragraph{Differentiable Renderer.} Given intrinsic camera parameters and hand meshes in the camera space, we use the differentiable rendering technique~\cite{KatoBMAMKG20} to project 3D estimations into the 2D image space. By applying various losses between the rendered mask of hand silhouette and the input binary image, we can integrate self-supervision into our end-to-end optimization process for constraining global mesh registration. Here we use the  PyTorch3D~\cite{RaviRNGLJG20} implementation of the differentiable renderer. The loss functions will be detailed later in the subsequent sections.

\paragraph{RefineNet.} In the previous stage, we only consider several pose-related principal components of MANO for parametrizing hands in a low-dimensional space. The lower degree of freedom may have the expressiveness problem that the finer details of hand gestures cannot be ideally reconstructed. Therefore, we incorporate a small convolutional neural network, called the RefineNet module~\footnote{Our RefineNet module is different from the dense prediction network proposed in \cite{LinLMSR20} despite the same name.}, into our model to deal with this issue. The network takes the concatenation of the input image and the projected silhouette as input, and it produces residual coordinate values that serve as the point-wise offsets to the previously constructed mesh. The final output of our model is the summation of the originally reconstructed hand mesh and the offset predicted by the RefineNet. The refined joints are then estimated by MANO's joint regressor given the refined mesh as its input.

\subsection{Loss Functions} \label{ssec:losses}
The complete objective function that we design for training our Mask2Hand model comprises three main loss terms: 
\begin{equation}
\label{eq:full_loss}
    \mathcal{L} = \mathcal{L}_\mathrm{pose} + \mathcal{L}_\mathrm{silhouette} + \mathcal{L}_\mathrm{mesh}\, ,
\end{equation}
where we have the pose loss $\mathcal{L}_\mathrm{pose}$ on the predicted joints, the silhouette loss $\mathcal{L}_\mathrm{silhouette}$ on the 2D projected silhouette of the hand, and the mesh loss $\mathcal{L}_\mathrm{mesh}$ on the reconstructed 3D mesh.

\subsubsection{Pose Loss.} \label{sssec:pose_loss}
The pose loss in \cref{eq:full_loss} further consists of two loss functions that use the unaligned and aligned joint positions to evaluate the preliminary and refined joint predictions by the squared L2-norm error:
\begin{equation}
    \mathcal{L}_\mathrm{pose} = \lambda_\mathrm{J} \mathcal{L}_\mathrm{J} + \lambda_\mathrm{alignJ} \mathcal{L}_\mathrm{alignJ} \,,
\end{equation}
where we have $\mathcal{L}_\mathrm{J}$ for the unaligned joint evaluation and $\mathcal{L}_\mathrm{alignJ}$ for the aligned joint evaluation, with $\lambda_\mathrm{J}$ and $\lambda_\mathrm{alignJ}$ as the weight factors. The 3D positions of the $K$ joints are represented as a $K$ by $3$ matrix, \ie, $J \in \mathbb{R}^{K\times 3}$.
The loss function $\mathcal{L}_\mathrm{J}$ of the unaligned joint evaluation is computed as
\begin{equation}
    \mathcal{L}_\mathrm{J} = \dfrac{1}{K} \sum_{i=1}^{K} \left\lVert J_i - \widehat{J}_i \right\rVert_2^2 + \dfrac{1}{K} \sum_{i=1}^{K} \left\lVert J_i - \widehat{J}^\mathrm{ref}_i \right\rVert_2^2 \,,
\end{equation}
$J_i$ is the ground-truth 3D position of joint $i$, $\widehat{J}_i$ is the preliminarily reconstructed position of joint $i$, and $\widehat{J}^\mathrm{ref}_i$ is the final prediction after refinement. 

For the aligned joint loss, the predicted joints are aligned with the ground truth using orthogonal Procrustes Analysis (PA)~\cite{Schonemann96}, and then the L2 loss is applied to the aligned and the ground-truth joints:
\begin{equation}
\label{eq:aligned_pose_loss}
    \begin{aligned}
    \mathcal{L}_\mathrm{alignJ} &= \dfrac{1}{K} \sum_{i=1}^{K} \left\lVert J_i - \widetilde{J}_i \right\rVert_2^2 + \dfrac{1}{K} \sum_{i=1}^{K} \left\lVert J_i - \widetilde{J}^\mathrm{ref}_i \right\rVert_2^2 \,,\\
    & \text{where} \;
    \widetilde{J} = \mathrm{PA}(J, \widehat{J}) \;\text{and}\; \widetilde{J}^\mathrm{ref} = \mathrm{PA}(J, \widehat{J}^\mathrm{ref}) \,.
    \end{aligned}
\end{equation}
The procedure of alignment $\mathrm{PA}(J, \widehat{J})$ can be described as follows.
\begin{enumerate}
    \item Translate the joints so that their mean lies at the origin:
    \begin{equation*}
        J = J - \frac{1}{K}\sum_{i=1}^{K} J_i\,, \quad \widehat{J} = \widehat{J} - \frac{1}{K}\sum_{i=1}^{K} \widehat{J}_i \,.
        \end{equation*}
    \item Divide the joints set by its Frobenius norm to remove the uniform scaling component: 
    \begin{equation*}
        J = {J}/{\lVert J \rVert_F}\,, \quad \widehat{J} = {\widehat{J}}/{\lVert \widehat{J} \rVert_F} \,.
        \end{equation*}
    \item Solve the orthogonal Procrustes problem: given matrices $J$ and $\widehat{J}$, find an orthogonal matrix $Q \in \mathbb{R}^{3\times 3}$ that maps $J$ most closely to $\widehat{J}$, \ie, $\min\limits_{Q} \lVert JQ - \widehat{J} \rVert_F$ subject to $Q^\top Q = I_{3\times 3}$. The solution consists of $Q = UV^\top$ and $s = \sum_i \sigma_i$, where $U \Sigma V^\top$ is the singular value decomposition of $J^\top \widehat{J}$ and $\sigma_i$'s are the singular values.
    \item $\widetilde{J} = \left(\widehat{J} Q^\top \times s \right) \times  \lVert J \rVert_F +  \left(\sum_{i=1}^K J_i\right)/K$, where the Procrustes output is scaled by the norm and shifted by the mean that we derive from $J$ in the first two steps.
\end{enumerate}

\subsubsection{Silhouette Loss.} \label{sssec:silhouette_loss}

For the module of differentiable rendering, the ideal result is that the 2D projected hand silhouette matches the input binary mask exactly. Hence, our silhouette loss is based on the binary cross-entropy (BCE) and the contour loss: 
\begin{equation}
    \begin{split}
        \mathcal{L}_\mathrm{silhouette} = \lambda_\mathrm{bce} \mathrm{BCE}(S, B) + \lambda_\mathrm{contour} \mathrm{ContourLoss}(S, B) \,,
    \end{split}
\end{equation}
where $S$ is the rendered silhouette and $B$ is the input binary mask.
Our contour loss is a differentiable estimation of Chamfer distance between two contours, calculated as a pixel-wise multiplication between the contour of rendered silhouette and the pre-computed distance field of the ground-truth silhouette's contour. The detailed steps for computing the Chamfer distance are summarized as follows:
\begin{enumerate}
    \item Compute the distance transform \cite{FelzenszwalbH12} for the contour $\partial B$ of the ground-truth binary mask $B$ in the dataset to get the distance field $\Psi_{\partial B}$.
    \item Apply differentiable binarization to the rendered silhouette image $S$. In our case, we convert a pixel value $x \in [0, 1]$ in $S$ to an approximately binary value as $\dfrac{1}{1 + e^{-100 \, (x - 0.5)}} \simeq \{0, 1\}$.
    \item Zero out the pixel values that are greater than the threshold 0.5 to deal with the noise near the wrist caused by PyTorch3D differentiable rendering. The result of Steps 2 \& 3 is a clean and nearly binarized silhouette $\overline{S}$.
    \item Apply Laplacian operator to $\overline{S}$ to get $\Delta \overline{S}$. The contour $\partial\overline{S}$ is then obtained by $\partial\overline{S}=\tanh(\max\{\Delta \overline{S}, 0\})$, which maps the values on the contour to near $1$ and all the rest to $0$.
    \item Do element-wise product between the resulting contour $\partial\overline{S}$ and the distance field $\Psi_{\partial B}$ to retrieve the corresponding distance values and then take the sum to get $\mathrm{ContourLoss}(S, B)$.
\end{enumerate}
All of the aforementioned computations are differentiable. Therefore, we can readily include the silhouette loss in our method, and end-to-end train the network.

\subsubsection{Mesh Loss.} \label{sssec:Mesh_loss}
We employ the mesh loss to measure the error in predicting the 3D vertices of the hand mesh. Like the pose loss, the mesh loss is also evaluated on both the unaligned and aligned predictions: 
\begin{equation}
    \mathcal{L}_\mathrm{mesh} = \lambda_\mathrm{V} \mathcal{L}_\mathrm{V} + \lambda_\mathrm{alignV} \mathcal{L}_\mathrm{alignV} \,.
\end{equation}
The unaligned vertex loss $\mathcal{L}_\mathrm{V}$ consists of the L1 losses for the predictions from the preliminary and refined meshes:
\begin{equation}
    \mathcal{L}_\mathrm{V} = \dfrac{1}{M} \sum_{i=1}^{M} \left\lVert V_i - \widehat{V}_i \right\rVert_1 + \dfrac{1}{M} \sum_{i=1}^{M} \left\lVert V_i - \widehat{V}^\mathrm{ref}_i \right\rVert_1 \,,
\end{equation}
where $V \in \mathbb{R}^{M\times 3}$ is the ground-truth 3D vertices of the hand mesh, $\widehat{V}$ is the preliminarily reconstructed mesh, and $\widehat{V}^\mathrm{ref}$ is the final prediction of vertices after refinement. \\
For the aligned vertex loss $\mathcal{L}_\mathrm{alignV}$, the predicted vertices are aligned with the ground truth using orthogonal Procrustes analysis as described in the aligned pose loss \cref{eq:aligned_pose_loss}, and then the L1 loss is applied to the aligned and the ground-truth mesh as
\begin{equation}
    \begin{aligned}
    \mathcal{L}_\mathrm{alignV} &= \dfrac{1}{M} \sum_{i=1}^{M} \left\lVert V_i - \widetilde{V}_i \right\rVert_1 + \dfrac{1}{M} \sum_{i=1}^{M} \left\lVert V_i - \widetilde{V}^\mathrm{ref}_i \right\rVert_1 \,, \\
    & \text{where} \;
    \widetilde{V} = \mathrm{PA}(V, \widehat{V}) \;\text{and}\; \widetilde{V}^\mathrm{ref} = \mathrm{PA}(V, \widehat{V}^\mathrm{ref}) \,.
    \end{aligned}
\end{equation}


\section{Experiments}

Since the problem formulation of the task that we aim to address is different from those commonly adopted in prior work, we have not found state-of-the-art methods that take exactly the same input setting and could therefore be considered for direct comparison with our approach. Note that the work of Lee \etal~\cite{LeeLCI19} is only posted on arXiv as a preprint without available code. Therefore, we do not include their method in the following comparisons. Moreover, their method requires the depth maps during training, which is more restrictive than ours. 
In our experiments, we first compare our method with the state-of-the-art depth-based model A2J~\cite{XiongZXCYZY19} on a synthetic dataset, and show that transferring our task for an existing model to solve is probably not as straightforward as it may seem. Further, we use the real data in FreiHAND to compare our method with the three state-of-the-art RGB-based methods: I2L-MeshNet~\cite{MoonL20}, 
MeshTransformer~\cite{LinWL21}, and
CMR (ResNet18)~\cite{ChenLMCWCGWZ21}. The results show that our method performs particularly well under the unaligned evaluation setting of 3D hand pose estimation, which is more practical for real applications.

\subsection{Datasets} \label{ssec:datasets}
We conduct experiments on the following two datasets.

\paragraph{Synthetic Dataset.} This dataset contains 20,000 training and 2,000 test binary images synthesized using MANO~\cite{RomeroTB17, TaheriGBT20}. For each sample in the dataset, its hand-pose principal components $\theta \in \mathbb{R}^{6}$ are randomly sampled from the uniform distribution between $[-2.0, 2.0)$, and each rotation angle of its 3D global orientation $R$ is randomly sampled from the uniform distribution $[-\pi, \pi)$. We use MANO to generate the corresponding hand shapes from the sampled parameters. The ground-truth 3D coordinates of the 21 hand joints can thus be automatically derived from the MANO rendered hand shapes.

\paragraph{FreiHAND~\cite{ZimmermannCYRAB19}.} It is a common real-world RGB dataset with 32,560 training samples and 3,960 test samples. We binarize the segmentation masks available in the training set and use them as the input to our model. However, since its original evaluation set does not provide segmentation masks, we resort to split its original training set into a random partition of 26,000 training, 3,280 validation, and 3,280 test images to achieve a reasonably fair evaluation on our method.

For quantitative evaluation, we report the standard metrics used by the FreiHAND dataset on the predictions of 3D joints and meshes: 
\begin{description}[itemsep=1pt,topsep=1pt]
\item \textbf{MPJPE}: the {\em mean per joint position error}, which measures the Euclidean distance (in cm) between the ground-truth joints and the predicted joints. 
\item \textbf{MPVPE}: the {\em mean per-vertex position error}, which measures the Euclidean distance (in cm) between the ground-truth vertices and the predicted vertices of a mesh. 
\item \textbf{AUC of PCK}: the {\em area under the curve} of the {\em percentage of correct keypoints}, which is plotted using 100 equally-spaced thresholds between 0 cm to 5 cm. 
\item \textbf{AUC of PCV}: the {\em area under the curve} of the {percentage of correct vertices}. 
\item \textbf{PA-MPJPE} or \textbf{PA-MPVPE}: It first applies Procrustes alignment between the ground truth and the prediction, and then calculates MPJPE or MPVPE. This metric aims to measure reconstruction error that neglects the effect of global rotation, translation, and scaling. 
\item \textbf{mIoU} and \textbf{Dice coefficient}: In addition to the preceding metrics used by the FreiHAND dataset for the evaluation of joint and mesh predictions, we also use the {\em mean intersection over union} and the {\em Dice coefficient} to evaluate the similarity between the input binary mask and the mask projected from the predicted hand mesh. 
\end{description}

\begin{table}[bth]
    \centering
    \caption{
        Comparison with the depth-based model A2J~\cite{XiongZXCYZY19} on the synthetic dataset. ({\bf Not Aligned})
    }
    \label{table:synthetic}
    \setlength{\tabcolsep}{3.25pt}
    \begin{tabular}{l|c|cc}
    \hline
    \multirow{1}{*}{Method} & \multirow{1}{*}{Input} & MPJPE (cm)$\downarrow$ & \multirow{1}{*}{AUC$_J$ $\uparrow$} \\
    \hline
    \multirow{2}{*}{A2J~\cite{XiongZXCYZY19}} & Mean depth + & \multirow{2}{*}{7.73} & \multirow{2}{*}{0.35} \\
     & 5-view augmentation &  &  \\
    \hline
    \multirow{2}{*}{Ours} & Binary (0, 255) + & \multirow{2}{*}{\textbf{0.87}} & \multirow{2}{*}{\textbf{0.83}} \\
     & 5-view augmentation &  &  \\
    \hline
    \end{tabular}
\end{table}

\begin{table}[bth]
    \centering
        \caption{
        Comparison with the depth-based model A2J~\cite{XiongZXCYZY19} on the synthetic dataset. ({\bf Procrustes Aligned})
    }
    \label{table:synthetic_aligned}
    \setlength{\tabcolsep}{3.25pt}
    \begin{tabular}{l|c|cc}
    \hline
    \multirow{1}{*}{Method} & \multirow{1}{*}{Input} & PA-MPJPE (cm)$\downarrow$ & \multirow{1}{*}{AUC$_J\uparrow$} \\
    \hline
    \multirow{2}{*}{A2J~\cite{XiongZXCYZY19}} & Mean depth + & \multirow{2}{*}{1.93} & \multirow{2}{*}{0.62} \\
     & 5-view augmentation &  &  \\
    \hline
    \multirow{2}{*}{Ours} & Binary (0, 255) + & \multirow{2}{*}{\textbf{0.53}} & \multirow{2}{*}{\textbf{0.89}} \\
     & 5-view augmentation &  &  \\
    \hline
    \end{tabular}
\end{table}

\begin{table*}[t]
    \centering
    \caption{
        FreiHAND ({\bf Not Aligned}). Note that the mIoU and the Dice coefficient are only relevant to the binary input case.
    }
    \label{table:freihand_not_aligned}
    \setlength{\tabcolsep}{3.25pt}
    \begin{tabular*}{\textwidth}{l|c|cccccc}
    \hline
    \multirow{2}{*}{Method} & \multirow{2}{*}{Input} & MPJPE & AUC of & MPVPE & AUC of & \multirow{2}{*}{mIoU$\uparrow$} & \multirow{2}{*}{Dice$\uparrow$} \\
     &  & (cm)$\downarrow$ & PCK$\uparrow$ & (cm)$\downarrow$ & PCV$\uparrow$ & &  \\
    \hline
    \multirow{2}{*}{MeshTransformer~\cite{LinWL21}} & RGB & 68.59 & 0.00 & 68.58 & 0.00 & -- & -- \\
     & Binary & 68.59 & 0.00 & 68.59 & 0.00 & 0.20 & 0.12 \\
    \hline
    \multirow{2}{*}{CMR (ResNet18)~\cite{ChenLMCWCGWZ21}} & RGB & 3.49 & 0.42 & 3.49 & 0.42 & -- & -- \\
     & Binary & 4.31 & 0.35 & 4.31 & 0.35 & \textbf{0.90} & \textbf{0.90} \\
    \hline
    Ours (6 pose PCs) & \multirow{2}{*}{Binary} & \textbf{3.56} & 0.40 & \textbf{3.57} & 0.40 & 0.88 & 0.87 \\
    Ours (45 pose PCs) &  & \textbf{3.56} & \textbf{0.41} & \textbf{3.57} & \textbf{0.41} & 0.88 & 0.88 \\
    \hline
    \end{tabular*}
\end{table*}

\begin{table*}[t]
    \centering
    \caption{
        FreiHAND ({\bf Procrustes Aligned}). The additional entries of PA-MPJPE and PA-MPVPE of I2L-MeshNet with RGB input are copied from the original paper~\cite{MoonL20} and posted here for the ease of reference, which can be considered as a strong baseline.
    }
    \label{table:freihand_aligned}
    \setlength{\tabcolsep}{3.25pt}
    \begin{tabular}{l|c|cccc}
    \hline
    \multirow{2}{*}{Method} & \multirow{2}{*}{Input} & PA-MPJPE & AUC of & PA-MPVPE & AUC of \\
     &  & (cm)$\downarrow$ & PCK$\uparrow$ & (cm)$\downarrow$ & PCV$\uparrow$ \\
    \hline
    \multirow{1}{*}{I2L-MeshNet~\cite{MoonL20}} & RGB & 0.74 & -- & 0.76 & --\\
    \hline
    \multirow{2}{*}{MeshTransformer~\cite{LinWL21}} & RGB & 0.54 & 0.89 & 0.58 & 0.88 \\
     & Binary & 0.60 & 0.88 & 0.64 & 0.87 \\
    \hline
    \multirow{2}{*}{CMR (ResNet18)~\cite{ChenLMCWCGWZ21}} & RGB & 0.67 & 0.87 & 0.68 & 0.87 \\
     & Binary & 0.77 & 0.85 & 0.77 & 0.85 \\
    \hline
    Ours (6 pose PCs) & \multirow{2}{*}{Binary} & 0.73 & 0.85 & 0.76 & 0.85 \\
    Ours (45 pose PCs) &  & 0.68 & 0.86 & 0.69 & 0.86 \\
    \hline
    \end{tabular}
\end{table*}

\begin{table*}[t]
    \centering
    \caption{
        Ablation Study. (Meaning of $\divideontimes$ in the first column: the aligned loss is only applied to the refined joints and vertices.)
    }
    \label{table:ablation_study}
    \begin{tabular}{ccccc|ccccc}
    \hline
    Aligned & BCE & Contour & Refine & Mesh & MPJPE & MPVPE & PA-MPJPE & PA-MPVPE & \multirow{2}{*}{mIoU$\uparrow$} \\
    Loss & Loss & Loss & Net & Loss & (cm)$\downarrow$ & (cm)$\downarrow$ & (cm)$\downarrow$ & (cm)$\downarrow$ &  \\
    \hline
    \checkmark &  & \checkmark & \checkmark & \checkmark & 3.55 & 3.56 & 0.78 & 0.79 & 0.87 \\
    \checkmark &  & \checkmark & \checkmark &  & 3.67 & 3.84 & 0.86 & 1.18 & 0.76 \\
    \checkmark &  & \checkmark &  &  & 3.73 & 3.73 & 0.94 & 0.94 & 0.83 \\
    \hline
    \checkmark &  &  & \checkmark & \checkmark & 3.60 & 3.61 & 0.80 & 0.81 & 0.87 \\
     &  & \checkmark & \checkmark & \checkmark & 3.83 & 3.91 & 1.36 & 1.56 & 0.86 \\
     $\divideontimes$ &  &  & \checkmark & \checkmark & 3.66 & 3.67 & 0.79 & 0.80 & 0.86 \\
    \hline
    \checkmark & \checkmark & \checkmark & \checkmark & \checkmark & 3.56 & 3.57 & 0.78 & 0.79 & 0.87 \\
    \checkmark & \checkmark & \checkmark &  &  & 3.72 & 3.73 & 0.90 & 0.93 & 0.83 \\
     & \checkmark & \checkmark & \checkmark & \checkmark & 3.66 & 3.67 & 1.04 & 1.06 & 0.88 \\
    \hline
    \end{tabular}
\end{table*}

\subsection{Comparison with State-of-the-Art Depth-based Methods} \label{ssec:comparison_depth}
To demonstrate that state-of-the-art depth-based models cannot be easily transferred to the task with binary input, we compare our model with A2J~\cite{XiongZXCYZY19} on the synthetic dataset. The input image of our model is binarized to $\{0, 255\}$ while that of A2J is binarized to 0 (as the background) and the mean depth of joints to treat the depth-based model fairly. This way, in spite of the fact that the input image is binary-valued, we still provide sufficient statistics of depth information to A2J. 
As shown in \Cref{table:synthetic}, for the unaligned setting, our model achieves MPJPE of 0.87 cm, and A2J reaches MPJPE of 7.73 cm. The results indicate that A2J fails to learn to localize 3D hand joints accurately when the input is binary, even if the reference mean depth is given. In contrast, our method has the promising ability to recover the positions of the joints in the camera space. In terms of MPJPE after Procrustes alignment, our approach also outperforms A2J by a large margin, as presented in \Cref{table:synthetic_aligned}. Some other depth-based models are not quantitatively compared with ours for the following reasons. V2V-PoseNet~\cite{MoonCL18} takes a 3D voxelized depth map as its input, which is expected to fail for silhouette-based input since the voxelization process cannot work for binary images. Hand PointNet~\cite{GeCWY18} converts a depth map into a 3D point cloud and then uses it as the model's input. Likewise, the conversion procedure cannot function properly for silhouettes, making it inadequate for the task of binary input.

\subsection{Comparison with State-of-the-Art RGB-based Methods} \label{ssec:comparison_rgb}
For the comparison with state-of-the-art RGB-based methods, we select the leading models from the FreiHAND competition leaderboard at CodaLab~\footnote{\url{https://competitions.codalab.org/competitions/21238}}, including CMR~\cite{ChenLMCWCGWZ21} and MeshTransformer~\cite{LinWL21}. To make a fair comparison with our method, we use the officially released code to re-train their models on the FreiHAND dataset randomly partitioned by us, as mentioned in \cref{ssec:datasets}, under the settings of both binary and RGB input. The former is for performance observation, while the latter is primarily for convincing proof that these models are not trained imperfectly. As presented in \Cref{table:freihand_not_aligned}, CMR achieves MPJPE and MPVPE of 4.31 cm with silhouette-based input, while MeshTransformer fails to localize 3D hand joints and mesh in the camera space. Our model reaches MPJPE of 3.56 cm and MPVPE of 3.57 cm, which outperforms the previous two methods by a large margin and is even comparable to the result of CMR with RGB input. The superior performance of our model in the unaligned case indicates that it has a better ability to reconstruct 3D hand mesh in the camera space, which is more practical in real applications. In terms of mIoU and Dice coefficient, our model achieves similar results as CMR. 

When the predicted joints and vertices are aligned with the ground truth by Procrustes analysis, our approach reaches PA-MPJPE of 0.68 cm and PA-MPVPE of 0.69 cm. These experimental results are significantly better than CMR's performance under the setting of binary input and are comparable to CMR's results with RGB input. Furthermore, unlike MeshTransformer, which totally fails in the unaligned case, our method performs stably well in both cases. Such a balanced behavior is preferable in our task since localization and detail recovery are vitally important for projecting a silhouette similar to the input one.

\subsection{Implementation Details} \label{ssec:implementation}
Our encoder's backbone is based on ResNet18~\cite{HeZRS16}. The PyTorch3D's implementation~\cite{RaviRNGLJG20} is employed for our differentiable renderer. We use the Adam optimizer to train our network with a batch size of 32. The initial learning rate is set to $10^{-4}$, and we use the ReduceOnPlateau scheduler for adjusting the learning rate. The whole network is trained for 150 epochs, and the checkpoint with the lowest validation loss is selected for our final usage. For the FreiHAND dataset, we apply data augmentation of random rotation between $[-\pi, \pi)$ and scaling between $[~0.9, 1.1)$ in the 2D image space. For the synthetic dataset, we randomly sample only one of the five views for each training data at every epoch to perform data augmentation. We set $\lambda_\mathrm{J} = 2 \times 10^{-3}$, $\lambda_\mathrm{alignJ} = 2 \times 10^{-2}$, $\lambda_\mathrm{contour} = 10^{-4}$, $\lambda_\mathrm{V} = 0.1$, and $\lambda_\mathrm{alignV} = 1$ to balance each loss term. $\lambda_{bce}$ is set to 0 for our final model and 0.5 for the ablation study.

\subsection{Ablation Study} \label{ssec:ablation}

We evaluate different configurations of our method to analyze the effectiveness of each component and examine the effects of the loss functions. 

\subsubsection{Effects of Shape Parameters.} In our ablation study, we use six pose PCs and set $\beta$ to the mean shape, \ie, $\beta = \bf{0}$. By comparing the first row of \Cref{table:ablation_study} and the penultimate row of \Cref{table:freihand_not_aligned,table:freihand_aligned}, we can observe that regressing shape parameters in the first stage of our model helps reduce the burden of refinement and achieve better performance in the aligned case.
 
\subsubsection{Effectiveness of Each Component.} 
As shown in \Cref{table:ablation_study}, we explore the evaluation results on the FreiHAND dataset with the exclusion of different components. Our base model consists of Encoder, MANO Layer, and Differentiable Renderer. After incorporating the RefineNet module into our network with pose loss only, we observe that both MPJPE and PA-MPJPE decrease while MPVPE and PA-MPVPE undesirably increase. This result is in line with our expectation since the pose loss cannot impose strong enough constraints on the deformation of the preliminarily constructed hand mesh. Based on the observation, we further apply the mesh loss to the output of the RefineNet, which helps to strengthen the functionality of refinement and improve all evaluation metrics significantly.

To reveal the effects of other losses, we exclude them one at a time from our entire network. First, we remove the contour loss between the input binary image and the projected silhouette. The evaluation result shows that the model's performance on both hand pose and shape estimation degrades, especially the unaligned ones, demonstrating the effectiveness of the contour loss for achieving more accurate global mesh registration. Second, we eliminate the aligned losses of both joints and vertices. Our experimental result shows that MPJPE, MPVPE, PA-MPJPE, and PA-MPVPE increase by 0.28, 0.35, 0.58, and 0.77, respectively. Such a significant deterioration in performance indicates that the aligned losses not only play crucial roles in the reconstruction of fine-grained gestures but also help to improve mesh recovery in the camera space. Finally, we remove all losses inserted before RefineNet. The result shown in the sixth row of \Cref{table:ablation_study} suggests that adding the pose loss and contour loss before the refinement stage can help our model improve the ability to localize the 3D hand mesh in the camera space.

\subsubsection{Effects of Adding the BCE Loss.} 
We also conduct some experiments to see whether adding the binary cross-entropy loss between the input binary mask and the projected silhouette has any positive effect. As listed in \Cref{table:ablation_study}, directly incorporating the BCE loss into our entire architecture contributes nothing to the final performance. To understand the root cause of this counterintuitive phenomenon, we take away the RefineNet module and the aligned loss one at a time with the presence of BCE loss. The experimental results show that the BCE loss can enhance our model's performance by a large margin when the aligned losses are absent. This implies that the BCE loss does not have much effect when the model learns the shape estimation task to a certain degree guided by the stronger training objectives, \ie, the aligned losses of joints and vertices.


\section{Conclusion}

This paper introduces the Mask2Hand network that learns to predict the 3D hand pose and shape from a 2D binary mask without relying on any RGB or depth information. Despite the more challenging input setting of our task in comparison with prior work on hand pose and shape estimation, we show that the proposed method can achieve a comparable performance as state-of-the-art RGB-based and depth-based methods. With the parametric hand model and the differentiable rendering technique, we integrate the self-supervised mechanism into our end-to-end training process without the need of human annotations. We propose several loss functions to model different aspects of 3D hand pose and shape estimation. Since our method explicitly learns to predict the global transformation of the hand, its better performance on the unaligned setting is thus promising for real applications.


\clearpage
%
%
\bibliographystyle{splncs04}
\bibliography{egbib}

\clearpage

\begin{center}
    \Large\bfseries\boldmath
    Supplementary Material
\end{center}

\setcounter{table}{0}
\renewcommand{\thetable}{A\arabic{table}}
\setcounter{figure}{0}
\renewcommand{\thefigure}{A\arabic{figure}}
\renewcommand{\theHtable}{Supp.\thetable}
\renewcommand{\theHfigure}{Supp.\thefigure}

\section*{A. Encoder}
The detailed model architecture for our Encoder network is illustrated in \Cref{table:encoder}. The feature extractor consists of a ResNet18 backbone and one fully-connected layer. Its output is fed into four distinct MLPs to produce parameters for the MANO layer. 

\begin{table}[bth]
    \centering
    \caption{
        Architecture for Encoder.
    }
    \label{table:encoder}
    \begin{tabular}{@{}c@{ }|@{ }c@{ }c}
    \hline
    Component & Layer & Output \\
    \hline
    \multirow{2}{*}{\begin{tabular}{@{}c@{}}Feature Extractor\\ \end{tabular}}
    & ResNet18 & $1 \times 512$ \\
     & Linear(512, 512), ReLU & $1 \times 512$ \\
    \hline
    \multirow{2}{*}{\begin{tabular}{@{}c@{}}Pose PCs Estimator\\ \end{tabular}}
    & Linear(512, 256), ReLU & $1 \times 256$ \\
     & Linear(256, \#PCs) & $1 \times \text{\#PCs}$ \\
    \hline
    \multirow{3}{*}{\begin{tabular}{@{}c@{}}Rotation Estimator\\ \end{tabular}} & Linear(512, 256), ReLU & $1 \times 256$ \\
     & Linear(256, 128), ReLU & $1 \times 128$ \\
     & Linear(128, 3) & $1 \times 3$ \\
    \hline
    \multirow{3}{*}{\begin{tabular}{@{}c@{}}Translation Estimator \\ \end{tabular}} & Linear(514, 256), ReLU & $1 \times 256$ \\
     & Linear(256, 128), ReLU & $1 \times 128$ \\
     & Linear(128, 3) & $1 \times 3$ \\
    \hline
    \multirow{2}{*}{\begin{tabular}{@{}c@{}}Shape Estimator\\ \end{tabular}}
    & Linear(512, 256), ReLU & $1 \times 256$ \\
     & Linear(256, 10) & $1 \times 10$ \\
    \hline
    \end{tabular}
    \vspace{-1em}
\end{table}


\setcounter{table}{0}
\renewcommand{\thetable}{B\arabic{table}}
\setcounter{figure}{0}
\renewcommand{\thefigure}{B\arabic{figure}}

\section*{B. More Qualitative Results}
We show more qualitative results on the test data. Note that the pictures illustrated in the main paper and this section demonstrate the results of the unaligned case. That is, we do not perform Procrustes alignment for visualization.

\begin{figure*}[t]
    \centering
    \includegraphics[width=0.95\textwidth]{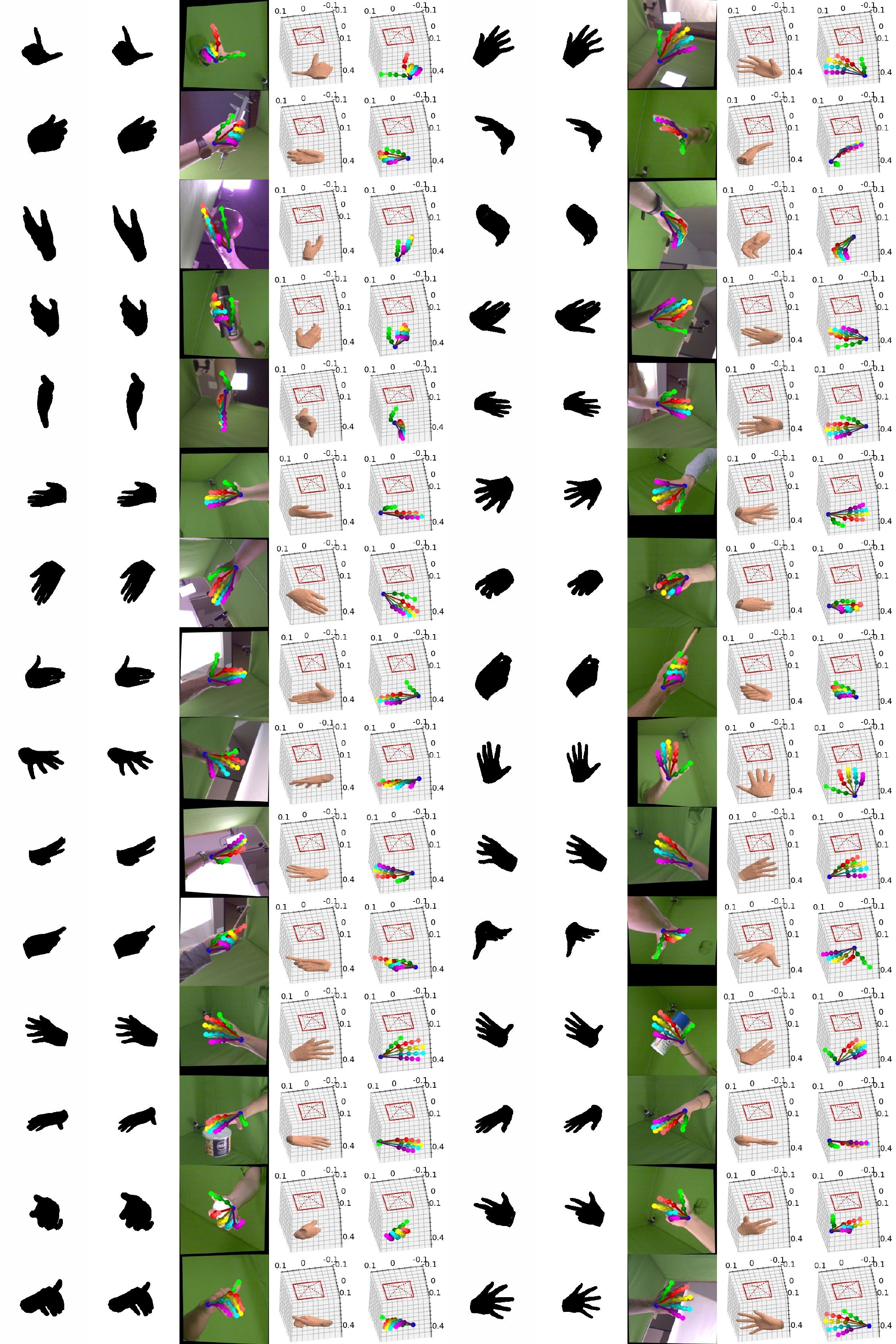}
    \vspace{-1em}
    \caption{
        Each result set contains a 2D binary mask as the input, followed by the predicted mask, 2D skeleton, mesh, and 3D skeleton.
    }
    \label{fig:supp_demo}
\end{figure*}

\begin{figure*}[t]
    \centering
    \includegraphics[width=0.95\textwidth]{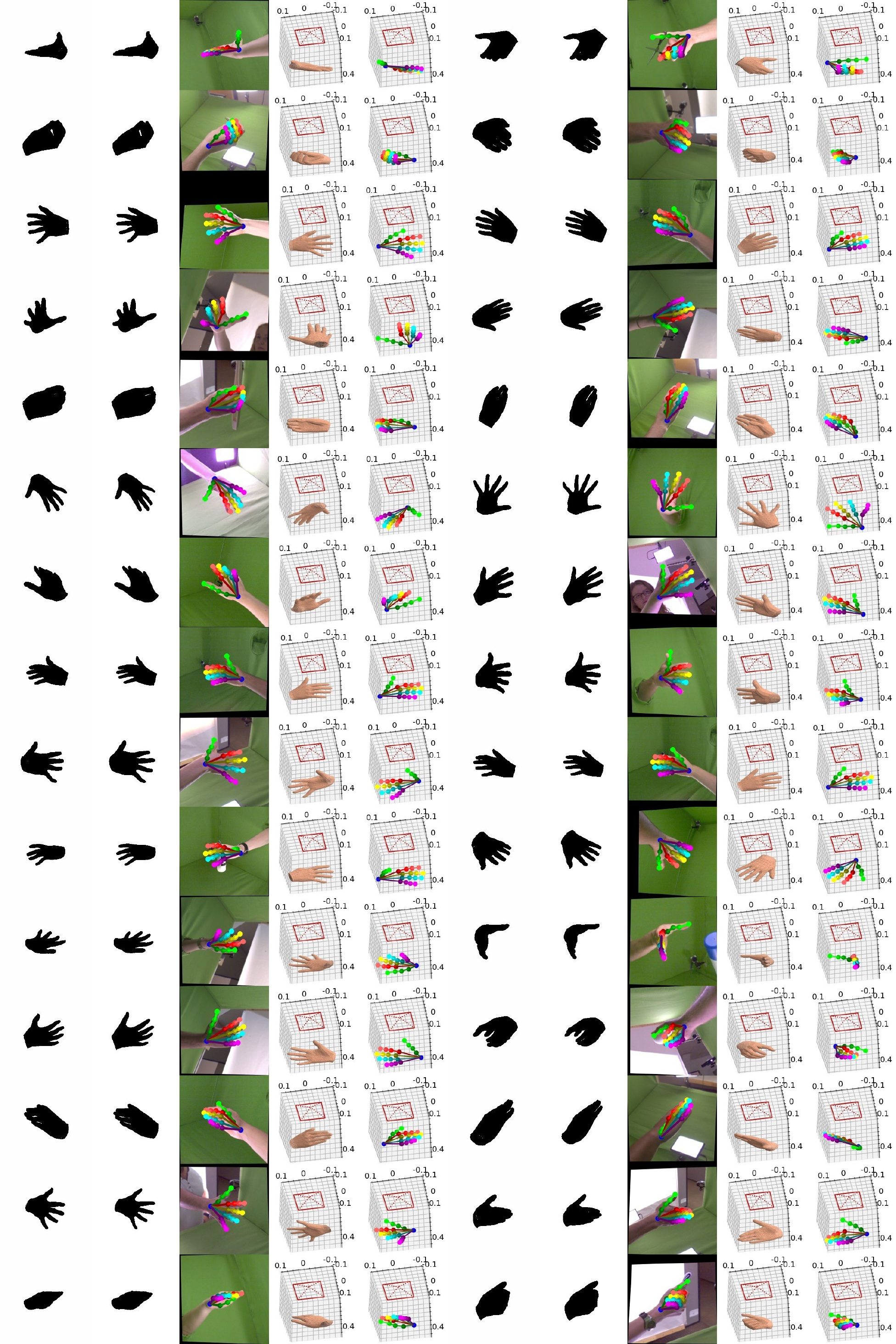}
    \vspace{-1em}
    \caption{
        Each result set contains a 2D binary mask as the input, followed by the predicted mask, 2D skeleton, mesh, and 3D skeleton.
    }
    \label{fig:supp_demo_2}
\end{figure*}

In \Cref{fig:supp_demo,fig:supp_demo_2}, we show two columns of 15 sets of results side by side. Each set contains five images: the first image presents the input of the 2D binary mask. The second image shows the rendered silhouette using the predicted hand shape. The third image shows the predicted 2D skeleton projected onto the image space. Note that the color images shown here are merely for visualization; our method does not require any RGB data. The last two images depict the predicted mesh and 3D skeleton in the camera space.

\setcounter{table}{0}
\renewcommand{\thetable}{C\arabic{table}}
\setcounter{figure}{0}
\renewcommand{\thefigure}{C\arabic{figure}}

\section*{C. Quantitative Results on Noisy Silhouettes}
\begin{figure*}[bth]
    \centering
    \includegraphics[width=\textwidth]{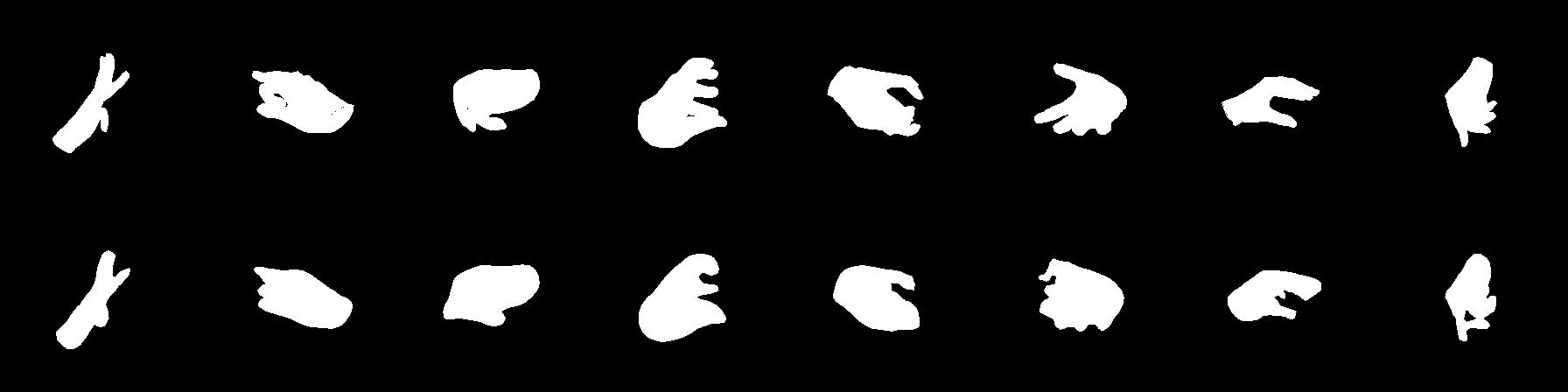}
    \vspace{-1em}
    \caption{
        Some bad results of the Deeplabv3 segmentation model. The first row presents the ground-truth masks, and the second row shows the predicted silhouettes. We can observe that finger poses in these examples are not reconstructed well, which can be considered strong noises injected into the inputs of hand pose estimation models.
    }
    \label{fig:deeplabv3_bad}
\end{figure*}

\begin{figure*}[bth]
    \centering
    \includegraphics[width=\textwidth]{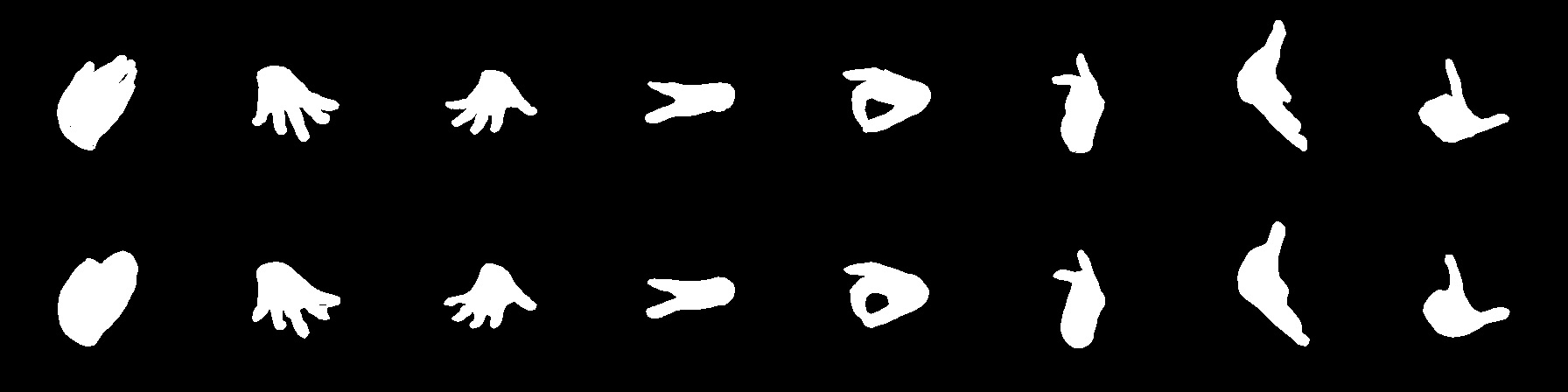}
    \vspace{-1em}
    \caption{
        Some good results of the Deeplabv3 segmentation model. The first row presents the ground-truth masks, and the second row shows the predicted silhouettes.
    }
    \label{fig:deeplabv3_good}
\end{figure*}

As relatively clean masks are used for experiments in the main paper, we aim to explore the robustness of each model under the setting of noisy silhouettes in this section. To achieve this goal, we first train a Deeplabv3 with ResNet-50 backbone as the hand segmentation model that takes an RGB image of a hand as input and produces its estimated mask. Then, we utilize the estimated masks (\ie noisy binary silhouettes) as inputs to train and evaluate both CMR and our method. Note that the silhouette loss is also calculated using the estimated masks in this experiment. The Deeplabv3 model reaches mIoU of 0.956 on the training data and 0.936 on the test data after being trained for seven epochs with a learning rate of $10^{-4}$. Despite the high mIoUs of Deeplabv3 predictions, by manually inspecting the produced silhouettes (see \Cref{fig:deeplabv3_bad,fig:deeplabv3_good}), we discover that finger poses in them are not reconstructed well, which can be considered a strong enough noise injected into the input of target models. As shown in \Cref{table:freihand_not_aligned_noisy,table:freihand_aligned_noisy}, our method still outperforms CMR by a large margin in all evaluation metrics. Further, the performance degradation of our model is significantly less than that of CMR in both unaligned and aligned cases, demonstrating that our approach is more robust to noisy input and thus more suitable for real-world applications.

\begin{table*}[bth]
    \centering
    \caption{
        FreiHAND ({\bf Not Aligned}).
    }
    \label{table:freihand_not_aligned_noisy}
    \setlength{\tabcolsep}{3.25pt}
    \begin{tabular}{l|c|cccc}
    \hline
    \multirow{2}{*}{Method} & \multirow{2}{*}{Input} & MPJPE & AUC of & MPVPE & AUC of \\
     &  & (cm)$\downarrow$ & PCK$\uparrow$ & (cm)$\downarrow$ & PCV$\uparrow$ \\
    \hline
    \multirow{2}{*}{CMR (ResNet18)} & Noisy binary & \multirow{2}{*}{6.32} & \multirow{2}{*}{0.24} & \multirow{2}{*}{6.33} & \multirow{2}{*}{0.24} \\
     & silhouette &  &  &  & \\
    \hline
    \multirow{2}{*}{Ours (45 PCs)} & Noisy binary & \multirow{2}{*}{\bf 4.44} & \multirow{2}{*}{\bf 0.32} & \multirow{2}{*}{\bf 4.47} & \multirow{2}{*}{\bf 0.32} \\
     & silhouette &  &  &  & \\
    \hline
    \end{tabular}
\end{table*}

\begin{table*}[bth]
    \centering
    \caption{
        FreiHAND ({\bf Procrustes Aligned}).
    }
    \label{table:freihand_aligned_noisy}
    \setlength{\tabcolsep}{3.25pt}
    \begin{tabular}{l|c|cccc}
    \hline
    \multirow{2}{*}{Method} & \multirow{2}{*}{Input} & PA-MPJPE & AUC of & PA-MPVPE & AUC of \\
     &  & (cm)$\downarrow$ & PCK$\uparrow$ & (cm)$\downarrow$ & PCV$\uparrow$ \\
    \hline
    \multirow{2}{*}{CMR (ResNet18)} & Noisy binary & \multirow{2}{*}{1.02} & \multirow{2}{*}{0.80} & \multirow{2}{*}{1.04} & \multirow{2}{*}{0.79} \\
     & silhouette &  &  &  & \\
    \hline
    \multirow{2}{*}{Ours (45 PCs)} & Noisy binary & \multirow{2}{*}{\bf 0.85} & \multirow{2}{*}{\bf 0.83} & \multirow{2}{*}{\bf 0.86} & \multirow{2}{*}{\bf 0.83} \\
     & silhouette &  &  &  & \\
    \hline
    \end{tabular}
\end{table*}

\setcounter{table}{0}
\renewcommand{\thetable}{D\arabic{table}}
\setcounter{figure}{0}
\renewcommand{\thefigure}{D\arabic{figure}}

\section*{D. Limitations}
\begin{figure*}[t]
    \centering
    \includegraphics[width=0.75\textwidth]{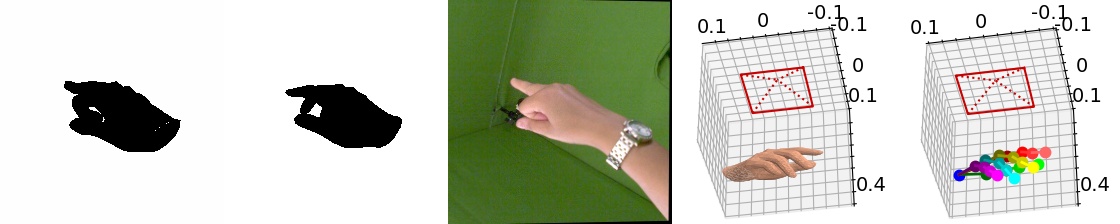}
    \vspace{-1em}
    \caption{
        A failure case. The leftmost image is an input 2D binary mask, followed by the predicted silhouette, ground-truth pose, reconstructed mesh, and 3D skeleton.
    }
    \label{fig:limitation_fig}
\end{figure*}

When there is apparent ambiguity in the input silhouette, our method might predict a distinct gesture that gives a very similar projected shadow. For instance, in the ground-truth image of \Cref{fig:limitation_fig}, the middle finger is extended, and the index finger is crooked. Our model predicts inversely, but the resulting silhouette is similar to the ground truth. This issue is quite challenging since the information of gradients in texture is lost in binary images.

\setcounter{table}{0}
\renewcommand{\thetable}{E\arabic{table}}
\setcounter{figure}{0}
\renewcommand{\thefigure}{E\arabic{figure}}

\section*{E. Clarification on Silhouette-Net}
We further explain why it is impossible to compare our method against Silhouette-Net on the HIM2017 dataset. HIM2017's depth images include human bodies and the captured hands are relatively small. The preprocessing step of this dataset requires cropping the hand segments by the given bounding boxes and resizing, which may lead to an unknown transformation between the camera space and the image space. The given camera intrinsics are thus inconsistent and misleading. Such an issue may be one of the reasons that \emph{depth supervision} or \emph{multi-view} is essential for a model (including Silhouette-Net) to succeed in HIM2017. Therefore, we consider HIM2017 inadequate for our task with \emph{single-view} binary information.


\clearpage

\end{document}